# Patient-Specific Game-Based Transfer Method for Parkinson's Disease Severity Prediction

Zaifa Xue, Huibin Lu, Tao Zhang, and Max A. Little

*Abstract*—Dysphonia is one of the early symptoms of Parkinson's disease (PD). Most existing methods use feature selection methods to find the optimal subset of voice features for all PD patients. Few have considered the heterogeneity between patients, which implies the need to provide specific prediction models for different patients. However, building the specific model faces the challenge of small sample size, which makes it lack generalization ability. Instance transfer is an effective way to solve this problem. Therefore, this paper proposes a patient-specific game-based transfer (PSGT) method for PD severity prediction. First, a selection mechanism is used to select PD patients with similar disease trends to the target patient from the source domain, which greatly reduces the risk of negative transfer. Then, the contribution of the transferred subjects and their instances to the disease estimation of the target subject is fairly evaluated by the Shapley value, which improves the interpretability of the method. Next, the proportion of valid instances in the transferred subjects is determined, and the instances with higher contribution are transferred to further reduce the difference between the transferred instance subset and the target subject. Finally, the selected subset of instances is added to the training set of the target subject, and the extended data is fed into the random forest to improve the performance of the method. Parkinson's telemonitoring dataset is used to evaluate the feasibility and effectiveness. Experiment results show that the PSGT has better performance in both prediction error and stability over compared methods.

## I. Introduction

PARKINSON'S disease (PD) is one of the common neurodegenerative diseases [1]. In addition to motor symptoms such as tremor, rigidity and muscle loss of control, its symptoms will also be manifested as dysphonia caused by lesions of the vocal system [2]. Studies have shown that dysphonia is one of the most common symptoms in the early stages of PD and can be detected up to 5 years before clinical diagnosis [3]. Moreover, there is strong evidence that PD progression is associated with worsening voice performance [4]. Therefore, as a non-invasive treatment method, voice pathology information of PD patients is often used to diagnose the severity of PD [5].

For the prediction of PD severity, scholars have worked to investigate the relationship between patients' voice features and symptom severity [6]. At present, the Unified Parkinson's Disease Rating Scale (UPDRS) is commonly used internationally to quantify the severity of PD [7]. Specifically, the UPDRS score is divided into motor-UPDRS and total-UPDRS, which describe the severity of the patients' motor symptoms and overall symptoms respectively. In exploring the relationship between patients' voice features and UPDRS, scholars initially have used all extracted voice features to predict the severity of PD patients' disease. Although these methods are simple and feasible, the prediction performance is poor [8]. Then, feature selection methods are used to remove irrelevant and redundant PD voice features to select appropriate feature subsets, thus improving the prediction performance of the model [9]-[11]. These models select the optimal subset of voice features for all PD patients overall.

However, building accurate predictive models between the voice feature set and symptom severity in PD patients is not straightforward because patients may have high heterogeneous in terms of demographics, genetic risk factors, comorbidity and staging [4], [12]. Currently, remote based home testing devices are increasingly being used to measure voice information of PD patients [13]. This convenient and cost-effective way enables patients to frequently assess PD severity and adjust their treatment plan in time to achieve optimal results. In the context of remote monitoring, patient heterogeneity means that each patient may have a different relationship between the voice features and the severity of PD, thus requiring the establishment of different prediction models [14]. However, building a specific predictive model for each patient faces the challenge of small sample size, making the model lack generalization ability [15].

Instance-based transfer learning can make up for the lack of training data of target subject by using the data of other patients [16], [17]. Transfer learning methods can be divided into four types: the instance transfer, feature representation transfer, parameter transfer, and relational knowledge transfer. Among them, the key idea of the instance transfer learning method is to select the most useful instances from the source

This work is supported by the National Natural Science Foundation of China under Grant 62176229 and the Natural Science Foundation of Hebei Province under Grant F2020203010 and Hebei Key Laboratory Project under Grant 202250701010046. (*Corresponding author: Tao Zhang; Max A. Little*).

Zaifa Xue is with the School of Information Science and Engineering, Yanshan University, Hebei Key Laboratory of Information Transmission and Signal Processing, Qinhuangdao, 066004, China (e-mail: xuezf@stumail.ysu.edu.cn).
Huibin Lu is with the School of Information Science and Engineering, Yanshan University, Hebei Key Laboratory of Information Transmission and Signal Processing, Qinhuangdao, 066004, China (e-mail: yjsbl@ysu.edu.cn).
Tao Zhang is with the School of Information Science and Engineering, Yanshan University, Hebei Key Laboratory of Information Transmission and Signal Processing, Qinhuangdao, 066004, China (e-mail: zhtao@ysu.edu.cn).
Max A. Little is with the School of Computer Science, University of Birmingham, Birmingham B15 2TT, U.K., and also with the Media Lab, MIT, Cambridge, MA 02139 USA (e-mail: maxl@mit.edu).

domain as additional training data to help predict the labels of the target subjects [18]. It is shown that instance transfer can be a promising tool to train a high-accuracy model with a small sample of labeled condition data, which is very effective in promoting the model performance [16]. Therefore, this paper analyzes the differences of disease progression among PD patients and increases the training data of the target subject by the instance transfer learning method to enhance the performance of the prediction model built for the target patient.

Most of the existing transfer learning methods for PD severity prediction transfer subjects from the source domain with similar disease progression between the target patient. However, not all the instances in the transferred subjects have a great positive impact on the disease prediction of the target patient. Therefore, this paper not only transfers subjects with similar disease development trends to the target patient, but also evaluates the contribution of instances in the transferred subjects, which further improves the effectiveness of the transferred instance subset. In cooperative game theory, the Shapley value is a method that can fairly assess the contribution of participants and satisfy the interpretability needs of the model [19], [20]. Inspired by this, we use the Shapley value to determine the contribution of PD subjects and their instances in the source domain to the prediction of PD severity in the target patient, so as to select the subset of instances with the high contribution and enhance the interpretability of the instance transfer process. Finally, the subset of instances with high contribution is considered important and transferred to the training set of the target patient, and the random forest (RF) method is used for prediction to enhance the generalization ability of the method.

By correlating and combining the above modules, we propose a patient-specific game-based transfer (PSGT) method for PD severity prediction. PSGT is a novel method to improve the PD severity prediction performance and the method interpretability. To verify the feasibility and effectiveness of the proposed method, the prediction experiment of the PSGT method is carried out on the real public Parkinson's telemonitoring dataset. The experiment results show that the PSGT method can select instances from the source domain that have small differences in disease progression and high contribution for the target patient, which improves the transfer robustness and achieves higher prediction accuracy. The main contributions of this work are four-fold:

1) The subject transfer mechanism is used to select subjects from the source domain with small differences in the disease development trend of the target subject, which greatly reduces the scope of instance transfer.
2) The importance of the transferred subjects is fairly evaluated by the Shapley value, and the proportion of valid instances in the transferred subjects is assigned accordingly.
3) The contribution of the instances in the transferred subjects is quantified using the simplified Shapley value to reduce the computational burden, and by combining with the subject's importance, the subset of instances with a high contribution to the disease prediction of the target patient is selected, which improves the performance and interpretability of the method.
4) The feasibility and effectiveness of the proposed method are verified by experiments on the real public Parkinson's telemonitoring dataset.

## II. RELATED WORK

### A. Prediction models based on machine learning

For the prediction of PD severity, scholars have proposed a variety of machine learning methods [13], [17], which may help doctors make decisions about the severity of patients [21].

The literature [8], [22]-[24] have used support vector regression (SVR), linear regression, k-nearest neighbor regression, and classification regression tree methods to predict the symptom severity of PD patients, respectively. These methods are simple, feasible, and fast to model and predict. However, there is a complex nonlinear relationship between the voice features of PD patients and UPDRS scores [25], and the errors obtained using a simple prediction model are large. To improve the accuracy of the prediction model, some neural network methods are proposed to address the severity of PD. Feed forward neural networks [26], [27] and adaptive network-based fuzzy information systems [28] have been used to predict patient's UPDRS due to their powerful nonlinear mapping capability. However, neural networks need to train a large amount of data and have high requirements for parameter settings. Usually, medical datasets are small and difficult to obtain. In this context, using neural network models to predict UPDRS in PD patients is prone to problems such as overfitting [29].

To take into account the model prediction performance and generalization ability, Tunc et al. [9] used the Boruta algorithm to select the most informative voice features, and combined it with the extreme gradient boosting (XGBoost) method to predict the UPDRS of patients, which improved the model generalization ability and provided a better understanding of the high-dimensional voice feature set. Tsanas et al. [10] combined feature selection methods with SVR, XGBoost, and RF, respectively. The results showed that RF always outperformed SVR and XGBoost in predicting PD severity. Therefore, this paper also uses RF to predict UPDRS scores in PD patients.

### B. Transfer learning method

Researchers have made some progress in using voice features to predict the severity of PD patients. However, the tested PD patients may have high heterogeneity. Therefore, it is necessary to develop specific predictive models for different PD patients. But there may be challenges with insufficient data when establishing prediction models for the target patient.

Transfer learning method can select data from the source domain that are similar to the target subject to increase the training data of the target patient [30], which is an effective method to solve the above challenges. Ji et al. [5] proposed a filtered domain adaptive model fusion (FDAMF) method, which filtered the source domain subject data that had a large difference in the trend of disease development from the target subject and constructed an accurate PD prediction model with the remaining subject data. Yoon et al. [4] used the positive transfer learning (PTL) method to predict the severity of target patients. Blind transfer will lead to the generation of negative transfer, so the PTL method also conducts theoretical research on the risk and conditions of negative transfer, which improves the robustness of transfer and the accuracy of the model. The transfer learning approaches proposed in [4] and [5] are committed to selecting subject data from the source domain that has little difference from the target subject for transferring. Although they have considered the differences of patients' diseases, they ignored the fact that the instances in the transferred subjects do not have the same contribution to the PD severity prediction of the target subject. In the modeling process of disease severity estimation of the target subject, selecting too much instances with small contribution may not effectively improve the prediction performance of the model. The transfer learning framework TrAdaBoost [31] utilizes the Boosting principle to automatically adjust the weights of the instances in the iterative process, which can eliminate the source domain instances that are not similar to the distribution of the target domain instances [32]. However, TrAdaBoost will lose its effectiveness when the difference in distribution between the source and target domains is too large [33]. This condition also limits the application of the TrAdaBoost algorithm to the prediction of disease severity in PD patients.

*C. The Shapley value*

To select a subset of instances with high importance from the source domain, it is also necessary to accurately assess the contribution of different instances in the transferred subjects, in addition to selecting subjects with similar disease development trends to the target subject. Inspired by the idea of cooperative game theory, we regard instance transfer in the transfer learning model as a game process to select a better subset of instances. The goal of cooperative game theory is to allocate different payoffs to game individuals in a fair and reasonable way. The Shapley value is a well-known solution concept in cooperative games [34], [35]. It is not only able to allocate the payoffs of game individuals in a fair way and thus get the importance of game individuals, but also meet the interpretability requirements of the model [36], [37]. Makarious et al. [38] assessed the importance of features associated with Parkinson's disease by using the Shapley value approach and created an accurate interpretation of each experiment result, thus enhancing the comprehensibility of the predictive model. Rahman et al. [39] extracted standard acoustic features from voice data and trained several machine learning algorithms to detect PD. In particular, they used the Shapley value to determine the importance of each feature to the model output to enhance the interpretability of the model, and concluded that the Shapley value could pick out the most important features in the predictive model.

In the PD severity assessment task, to select subjects or instances from the source domain that have similar disease progression to the target patient, each subject or instance can be treated as a participant, and the payoff is the difference between the predicted disease severity of the target patient before and after the transfer of subjects or instances. The analysis of the contribution of each subject or instance by the Shapley value reveals its influence on the prediction model and helps to enhance the interpretability of the PD severity prediction model. Therefore, combining the Shapley value with transfer learning for UPDRS prediction of PD patients is a way to improve the performance and interpretability of transfer model.

III. METHODOLOGY

In this section, each module of the proposed PSGT method is described in detail. The structure of the method is shown in Fig. 1. It includes two main modules: subject transfer and instance transfer. First, the similarity between each subject in the source domain and the target patient is evaluated, and a selection mechanism is used to select subjects with small differences in the disease development trend of the target patient. Next, the contribution of the transferred subjects and their instances is calculated by the Shapley value to further refine the importance of the instances in the transferred subjects to the disease prediction of the target patients. Finally, a subset of transferred instances is added to the training data of the target subject and RF is used to predict the PD severity of the target patient.

*A. Problem statement*

In this paper, the target subject is a PD patient arbitrarily selected from the dataset, and the rest of the patients are the subjects in the source domain (called source subjects). The input of the PSGT is the voice feature data of the source subject and the known target subject. The method effectively addresses the problem of limited target patient data by transferring subjects and their instances from the source domain to increase the training data of the target subject. Given the target patient's PD voice feature data $T$, it is divided into training set $T_{train}$, validation set $T_{val}$ and test set $T_{test}$. We aim to identify subjects and a subset of their instances from the source domain with small differences in PD development trend from the target patient. For this purpose, we use the training and validation set data of the target subject to evaluate the importance of the source subjects and instances. To reduce the scope of transfer, we first use the source subject data $S$ and the validation set data $T_{val}$ of the target subject to filter the subjects to be transferred, and select the top $k$ source subjects in importance:

$$S^T = f_{sel\_sub}(S, T_{val}, k) \quad (1)$$

where $f_{sel\_sub}(\cdot)$ is a function that selects source subjects. Then, we analyze the importance of each instance in the transferred subjects $S^T$ to find the subset of instances that contributed most to the disease prediction of the target subject. By adding the transferred instances to the training set $T_{train}$, a new training set

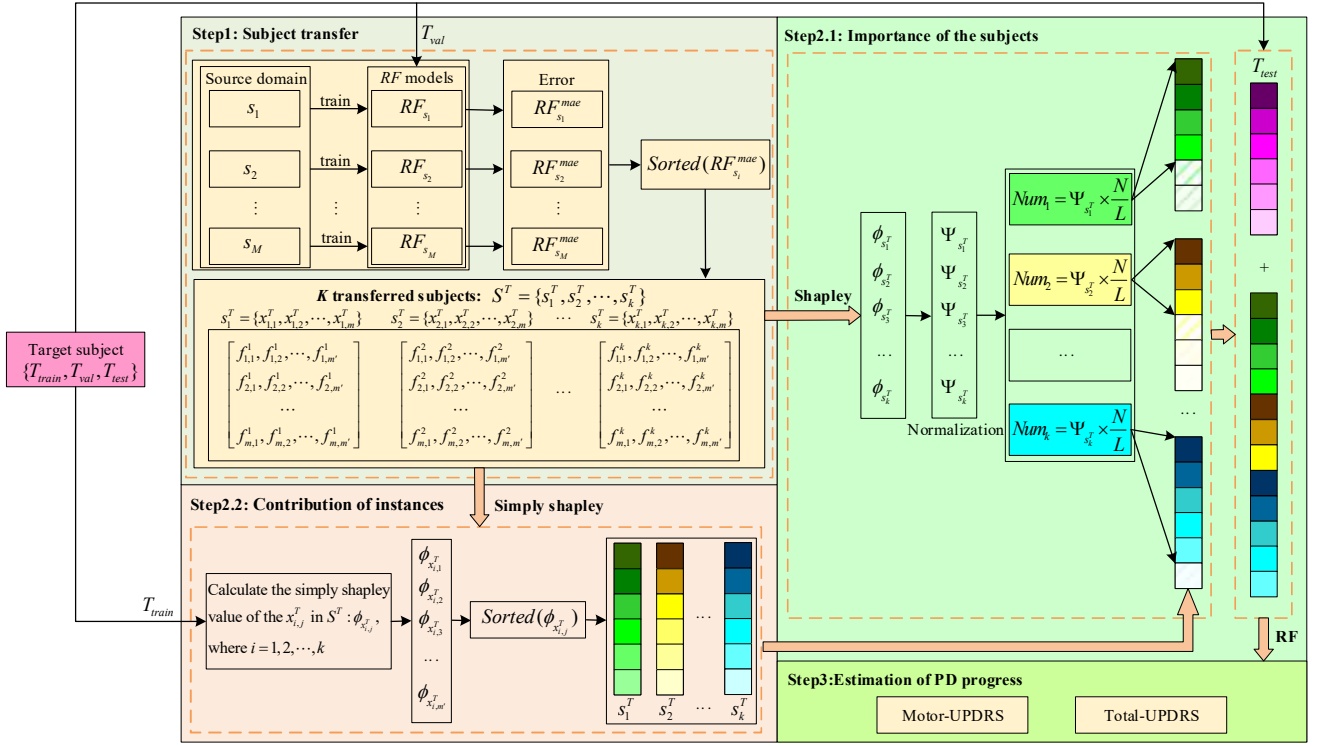

**Fig. 1.** Schematic of PSGT method for PD severity prediction.

$T'_{train}$ of the target subject is obtained:

$$T'_{train} = T_{train} \cup \{f_{sel\_spl}(S^T, T_{train}, T_{val})\} \quad (2)$$

where $f_{sel\_spl}(\cdot)$ is the function that selects the instances from the transferred subjects. To make accurate prediction of PD severity of the target subject, we input the training set $T'_{train}$ into a suitable nonlinear mapping function $f(\cdot)$, which can find out the relationship between the voice features of PD patients and the target value UPDRS. By testing it on $T_{test}$, the predicted value $y'$ of the target patient's disease can be obtained as follow:

$$y' = f(T'_{train}, T_{test}). \quad (3)$$

*B. Subject transfer*

The first part of the PSGT method is the subject transfer (ST). Fig. 2 illustrates the flow of subject transfer. In particular, the red curve in Fig. 2 (a) and (b) indicates the change of UPDRS of the target subject over time, and the curves of the other colors represent the disease development trend of the source subjects.

The input of ST module are the source subjects whose purpose is to train its corresponding PD severity prediction model. The validation set $T_{val}$ of the target subject is tested in the prediction model of each source subject. The difference of the disease development trend between each source subject and the target subject can be judged by the prediction error. This selection mechanism can select $k$ source subjects with high similarity to the disease of the target subject, which greatly reduces the scope of transfer instances and the risk of negative transfer.

The selection of an appropriate PD severity prediction model has an important impact on the analysis of disease variability among patients. As an effective ensemble learning algorithm, RF is widely used in the field of biomedical informatics [40]. Due to its high validity, stability and robustness [41], RF has also shown outstanding performance in PD symptom progression prediction [10], fitting well the nonlinear relationship between voice features and the target value UPDRS. Therefore, to select source subjects with similar disease trends to the target subject as shown in Fig. 2(b), this paper uses RF as a nonlinear mapping function to mine the intrinsic relationship between the PD voice features and the target value UPDRS. As shown in Fig. 2, the first step of the subject transfer model is to input all the instances of each subject $s_i$ in the source domain $S = \{s_1, s_2, \cdots, s_M\}$ into the RF, and its corresponding prediction model $RF(S)$ can be obtained by training:

$$RF(S) = \{RF_{s_1}, RF_{s_2}, \cdots, RF_{s_M}\}. \quad (4)$$

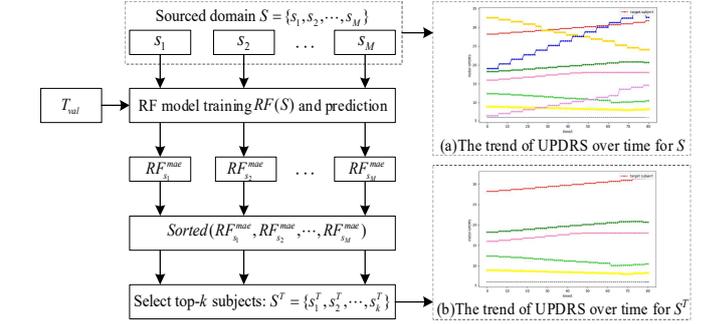

**Fig. 2.** Diagram of the flow of subject transfer.

To analyze the similarity of the disease development trend between the source subjects and the target subject, we input the validation set $T_{val}$ of the target subject into the prediction model $RF(S)$ of the source subject for testing. The difference between the source subject $s_i$ and the target subject is assessed by mean absolute error $RF_{s_i}^{mae}$:

$$RF_{s_i}^{mae} = \frac{1}{N_{val}} \sum_{i=1}^{N_{val}} |RF_{s_i}(T_{val}) - y_{val}| \qquad (5)$$

where $N_{val}$ is the number of instances in the validation set $T_{val}$. $y_{val}$ represents the true value of the patient's UPDRS in $T_{val}$. $RF_{s_i}(T_{val})$ and $RF_{s_i}^{mae}$ denote the predicted and mean absolute error values obtained by the RF model corresponding to the source subject $s_i$ to predict $T_{val}$, respectively. The proposed subject transfer method treats each source subject as a whole and trains its corresponding prediction model $RF(S)$. Therefore, the mean absolute error $RF_S^{mae} = \{RF_{s_1}^{mae}, RF_{s_2}^{mae}, \cdots, RF_{s_M}^{mae}\}$ obtained by $RF(S)$ predicting $T_{val}$ can be used to assess the variability between the source and target subjects. The smaller the $RF_{s_i}^{mae}$, the higher the similarity of the PD development trend between the source subject $s_i$ and the target subject. In this paper, the mean absolute error in $RF_S^{mae}$ is sorted from small to large by the function $sorted(\cdot)$, and the $k$ transferred subjects $S^T$ are identified:

$$S^T = S_{index_{1,\cdots,k}(sorted(RF_{s_1}^{mae}, RF_{s_2}^{mae}, \cdots, RF_{s_M}^{mae}))} \qquad (6)$$

where $S^T = \{s_1^T, s_2^T, \cdots, s_k^T\}$, $index_{1,\cdots,k}$ denotes the index of the top $k$ subjects selected with small mean absolute error.

*C. Instance transfer*

The subject transfer proposed in this paper selects $k$ subjects from the source domain with small differences in disease trends between them and the target patient, which greatly reduces the scope of instances to be transferred. However, the instances in the $k$ transferred subjects may contribute differently to the disease prediction of the target subject. Therefore, to select the subset of instances with highest contribution, it is necessary to fairly evaluate the importance of the instances so as to improve the prediction performance of the method effectively. Algorithm 1 demonstrates the instance transfer mechanism proposed in the second part of the PSGT method.

*1) Importance of the subjects:* To select a subset of instances with high contribution to the disease prediction of the target patient, the Shapley value is used in Algorithm 1 to evaluate the importance of the transferred PD patients and their instances. The Shapley value is a well-known solution in cooperative game theory. The cooperative game theory $\Gamma = (\mathcal{N}, \gamma)$ consists of a set of players $\mathcal{N} = \{1, 2, \cdots, n\}$ and a payoff function $\gamma(\mathcal{N})$. Where $\mathcal{N}$ is called the coalition and $\gamma(\mathcal{N})$ represents the payoff obtained by all players in $\mathcal{N}$ working together to complete the task. The core of cooperative game theory is to calculate how much gain is obtained and how to distribute the total gain fairly [35]. The Shapley value method proposed by Lloyd [42] not only distributes the benefits fairly but also assesses the importance of each player [19], [35]. The contribution of instances in the transferred subjects to the disease prediction of the target patient may not be the same, so the proportion of instances with high

---

**Algorithm 1** Instance Transfer Mechanism

**Input:** $S^T = \{s_1^T, s_2^T, \cdots, s_k^T\}$ : $k$ transferred subjects, where $s_i^T = \{x_{i,1}^T, x_{i,2}^T, \cdots, x_{i,m}^T\}$; $T_{train}$ and $T_{val}$ : the training set and the validation set of the target subject.

**Output:** $Spl$ : transferred subset of instances.

1: // Calculating the importance of the subjects.
2: **for** the subject $i \in \{1, 2, \cdots, k\}$ **do**
3:   Calculate the Shapley value of the $i^{th}$ subject $s_i^T$ : $\phi_{s_i^T}$.
4:   Map all Shapley values to positive values: $\phi'_{s_i^T}$.
5: **end for**
6: Normalize the importance of transferred subjects: $\Psi_{s_i^T}$.
7: // Calculating the importance of the instances.
8: **for** the subject $i \in \{1, 2, \cdots, k\}$ **do**
9:   **for** the instance $j \in \{1, 2, \cdots, m\}$ **do**
10:    Calculate the contribution of instance $x_{i,j}^T$ : $\phi_{x_{i,j}^T}$.
11:  **end for**
12: **end for**
13: // Identifying the instances to be transferred.
14: **for** the subject $i \in \{1, 2, \cdots, k\}$ **do**
15:   Rank the contribution of instances: $sorted(\phi_{x_{i,1}^T}, \phi_{x_{i,2}^T}, \cdots, \phi_{x_{i,m}^T})$.
16:   Determine the number of instances to transfer: $Num_i$.
17:   Select the instances: $s_i^T{}_{index_{1,\cdots,Num_i}(sorted(\phi_{x_{i,1}^T}, \phi_{x_{i,2}^T}, \cdots, \phi_{x_{i,m}^T}))}$.
18: **end for**
19: Merge transferred instances in $k$ subjects: $Spl$.

---

contribution in the $k$ transferred subjects may also be different. Inspired by the idea of cooperative game theory, we regard the evaluation of the importance of $k$ transferred subjects as a process of cooperative game among subjects, and use the Shapley value to analyze the importance of each subject. Specifically, the players in $\Gamma = (\mathcal{N}, \gamma)$ are replaced with $k$ transferred subjects, i.e., $\Gamma' = (S^T, \gamma)$. Then, the importance of the $k$ transferred subjects for predicting disease severity in the target patient can be calculated by using the Shapley value. From this, the proportion of instances in each subject that have a high contribution to the disease prediction of the target subject can be determined. Where the Shapley value $\phi_{s_i^T}$ for the subject $s_i^T$ can be expressed as:

$$\phi_{s_i^T} = \sum_{\mathcal{S} \subseteq S^T} \Delta_{s_i^T}(\mathcal{S}) \frac{|\mathcal{S}|!(k - \mathcal{S} - 1)!}{k!} \qquad (7)$$

$$\Delta_{s_i^T}(\mathcal{S}) = \gamma(\mathcal{S} \cup s_i^T) - \gamma(\mathcal{S}) \qquad (8)$$

where $\Delta_{s_i^T}(\mathcal{S})$ represents the marginal contribution of subject $s_i^T$. $\gamma(\mathcal{S})$ denotes the loss value obtained by the subcoalition $\mathcal{S}$ for predicting the disease severity of the target patient, and the smaller the value, the smaller the error of prediction using the subcoalition. Therefore, (8) is used to determine whether the subject $s_i^T$ can reduce the prediction error of $\mathcal{S}$ when subject $s_i^T$ joins the subcoalition $\mathcal{S}$. The Shapley value in (7) can be interpreted as the average of all marginal contributions of the subject $s_i^T$ [34]. The importance of subject $s_i^T$ increases if it can reduce the error of more subcoalitions $\mathcal{S}$. According to this rule, we evaluate the importance of the $k$ transferred subjects using the Shapley value and determine the proportion of instances with high contribution in each subject according to (9) and (10):

$$\phi'_{s_i^T} = 1 - sigmoid(\phi_{s_i^T}) = \frac{\exp(-\phi_{s_i^T})}{1+\exp(-\phi_{s_i^T}))} \quad (9)$$

$$\Psi_{s_i^T} = \frac{\phi'_{s_i^T}}{\varphi} = \frac{\phi'_{s_i^T}}{\sum_{i=1}^{k}\phi'_{s_i^T}} \quad (10)$$

where (9) transforms the Shapley value $\phi_{s_i^T}$ of subject $s_i^T$ to a positive value $\phi'_{s_i^T}$ by the sigmoid function, and keeps the order of the importance of the $k$ subjects unchanged. Equation (10) normalizes $\phi'_{s_i^T}$ to derive the proportion $\Psi_{s_i^T}$ of instances obtained from subject $s_i^T$. Therefore, by using (11), the number $Num_i$ of the transferred instances in the subject $s_i^T$ can be obtained, where $m$ represents the number of instances in the subject $s_i^T$:

$$Num_i = \Psi_{s_i^T} \times m. \quad (11)$$

*2) Contribution of instances:* In the proposed instance transfer method, in addition to determining the number of instances selected from the $k$ subjects $S^T = \{s_1^T, s_2^T, \cdots, s_k^T\}$, the contribution of the instance $s_i^T = \{x_{i,1}^T, x_{i,2}^T, \cdots, x_{i,m}^T\}$ in each subject needs to be evaluated. Where, $x_{i,m}^T = \{f_{m,l}^i\}_{l=1}^{m'}$ denotes the $m^{th}$ instance vector of the $i^{th}$ subject, $f$ denotes the $l^{th}$ feature of the $m^{th}$ instance, and $m'$ denotes the dimensionality of the instance vector. Similar to calculating the importance of the subject, the contribution of the instances can be calculated using the Shapley value:

$$\phi_{x_{i,j}^T} = \sum_{\mathcal{S} \subseteq s_i^T} \Delta_{x_{i,j}^T}(\mathcal{S}) \frac{|\mathcal{S}|!(m-\mathcal{S}-1)!}{m!} \quad (12)$$

where $\phi_{x_{i,j}^T}$ and $\Delta_{x_{i,j}^T}(\mathcal{S})$ denote the Shapley value and marginal contribution of the $j^{th}$ instance in the $i^{th}$ subject, respectively. However, the Shapley value must iterate through all possible coalitions for each instance. The number of instances is $m$, then the number of subcoalitions is $2^m$. Therefore, if the number of instances $m$ is large, using the original Shapley value method will have a heavy computational burden. To solve this problem, this paper uses the simplified Shapley value [43] to calculate the contribution of instances. This method is an effective variant of the original Shapley value, which greatly improves the computational efficiency while having interpretability.

The original Shapley value is obtained by calculating the mean of the marginal contributions of all possible subcoalitions for each PD patient. The subcoalition's contribution and marginal contribution can be measured in a variety of ways [43], [44]. Inspired by this, the simplified Shapley value replaces $\Delta_{x_{i,j}^T}(\mathcal{S})$ in (8) with the sum of $\gamma(\xi)$:

$$\Delta_{x_{i,j}^T}(\mathcal{S}) = \sum_{\xi:\{x_{i,j}^T\} \in \xi \wedge \xi \subseteq \mathcal{S} \cup \{x_{i,j}^T\}} \gamma(\xi), j=1,\cdots,m \quad (13)$$

where $\gamma(\xi)$ represents the error generated by the prediction after the instance set $\xi$ is input to the RF, which is called the individual contribution of the subcoalition $\xi$. Thus, the Shapley value of instance $x_{i,j}^T$ is the weighted average of the individual contributions of all possible coalitions $\xi$ including instance $x_{i,j}^T$. The simplified Shapley value is computed at most once for each subcoalition containing each instance without traversing all possible subcoalitions, which greatly improves the computational efficiency of the Shapley value. Eventually, (7) can be calculated using the following formula:

$$\phi_{x_{i,j}^T} = \sum_{\mathcal{S} \subseteq s_i^T \setminus \{x_{i,j}^T\}} \frac{1}{|\mathcal{S}|+1}\gamma(\mathcal{S} \cup \{x_{i,j}^T\}), j=1,\cdots,m \quad (14)$$

where $\gamma(\mathcal{S} \cup \{x_{i,j}^T\})$ represents the prediction error of all instances in the coalition $\mathcal{S} \cup \{x_{i,j}^T\}$. Combining the importance $\phi_{x_{i,j}^T}$ of instances and the number $Num_i$ of selected instances in the transferred subjects, the transferred subset $Spl$ of instances can be determined as follow:

$$Spl = \bigcup_{i=1}^{k} s_{i\ index_{1,\cdots,Num_i}}^T (sorted(\phi_{x_{i,1}^T}, \phi_{x_{i,2}^T}, \cdots, \phi_{x_{i,m}^T})). \quad (15)$$

*3) The final sample subset:* The subset of transferred instances $Spl$ is derived with the assumption that the number of instances of each source subject is the same by default, as shown in (11). However, in practical applications, this requirement may be difficult to satisfy. In addition, the literature [43] points out that when the number of transferred instances exceeds the number of instances of the target subject, it tends to lead to negative transfer, which may reduce the prediction accuracy of the model. To release the limitation of the number of subject instances on the model, we have optimized the number $Spl_{num}$ of transferred instances by further analyzing the structure of the proposed method.

Fig. 3 shows the flowchart of the PSGT method after optimization for the number $Spl_{num}$ of transferred instances. To avoid the number of transferred instances exceeding the target patient, we reduce the number of transferred instances in two steps to make it less than or equal to $N/L$ to reduce the risk of negative transfer [45]. Where $N$ denotes the number of instances in the target subject and $1/L$ denotes the ratio of the number of transferred instances to the total number of instances in the target subject. In the first step, we remove the instances with positive Shapley value. The contribution of the subcoalition in (14) is assessed by the prediction error, and a smaller Shapley value indicates a greater contribution of the instance to reduce the prediction error. Therefore, removing instances with large Shapley value is beneficial to reduce the negative impact of these instances on the disease prediction of the target patient. In addition, considering that in other practical application scenarios, the number of instances $m$ of source subjects may differ significantly from $N$. Due to the small sample size of PD patients, the number of transferred instances is likely to be much larger than the number of instances of the target subject. To reduce the interference of negative transfer and increase the generalizability of the PSGT method in different application scenarios, the number of transferred instances is set within $N/L$ in this paper.

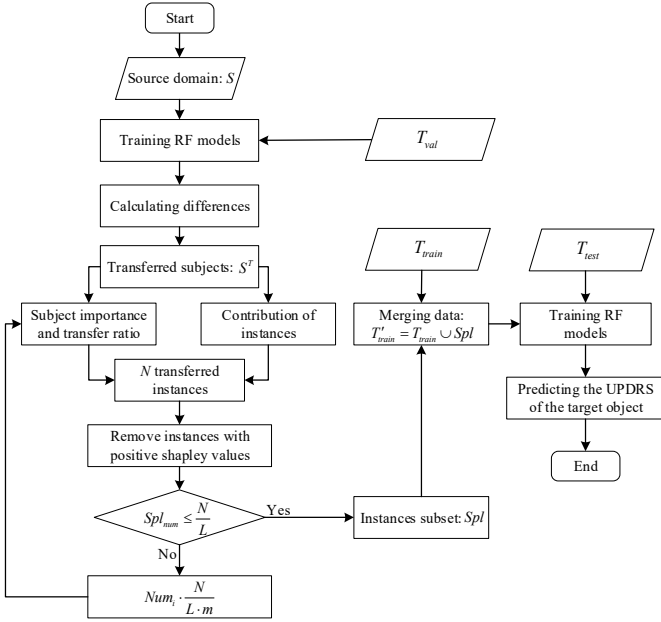

**Fig.3.** Flowchart of the PSGT method optimized for the number of transferred instances.

As shown in Fig. 3, the first stage of the proposed PSGT method is subject transfer, and $k$ subjects $S^T$ with less difference in the disease development trend from the target subject are obtained. In the second stage, the importance of $k$ subjects and their instances is evaluated using the Shapley value, so that the subset of instances $Spl$ with the high contribution to the disease prediction of the target patient is identified. Finally, the instance subset $Spl$ is added to the training set of the target subject to form a new training set $T'_{train}$, and the RF model is used to construct the nonlinear relationship between PD voice features and UPDRS. The proposed PSGT method takes into account the differences of diseases among different patients, selects the subjects with similar development trends to the target subject and instances with high contribution, and optimizes the number of transferred instances to solve the problem of limited data of the target patient. At the same time, it effectively reduces the risk of negative transfer of the method.

## IV. EXPERIMENTS

### A. Parkinson's telemonitoring dataset

To test the feasibility and validity of the PSGT method, we conduct prediction experiments using the Parkinson's telemonitoring dataset that is widely used for PD severity estimation [4], [5], [9]-[11], [31].

The Parkinson's telemonitoring dataset was collected by Tsanas et al. at the University of Oxford [8]. A total of 42 patients with PD were collected in this dataset, including 28 males and 14 females, and all patients had a 5-year history of the disease. First, PD patients were asked to record 6 sustained vowels /a/ on the same day of the week, where the patient recorded 4 sustained vowels /a/ at a comfortable pitch and loudness and the remaining 2 vowels /a/ were asked to be recorded at twice the loudness. The procedure lasted for 6 months and the patients were not allowed to take medication during the test, resulting in the collection of 5875 voice data. To facilitate the mapping between patients' voice and disease severity UPDRS, Tsanas et al. extracted 16 voice features from the raw voice data, called dysphonia measures, as shown in Table I. In addition, since gender and age are also important features in the diagnosis of PD [14], [46], they can be used in the assessment of the severity of PD symptoms.

TABLE I
VOICE FEATURES OF THE PARKINSON'S DISEASE
TELEMONITORING DATASET

| Classification | Voice features |
|---|---|
| Frequency disturbance | Jitter (%), Jitter (Abs), Jitter: RAP, Jitter: PPQ5, Jitter: DDP |
| Amplitude perturbation | Shimmer, Shimmer(dB), Shimmer: APQ3, Shimmer: APQ5, Shimmer: APQ11, Shimmer: DDA |
| Noise-signal ratio | NHR, HNR |
| Nonlinear fundamental frequency variation | RPDE, PPE |
| Nonlinear noise variation | DFA |

The severity score UPDRS of PD patients in the dataset was tested in 3 stages, and the UPDRS of each patient was assessed at the beginning of the test, after 3 months, and after 6 months, respectively. It has been shown that the UPDRS increases essentially linearly in early PD patients without medication [47]. Therefore, this dataset was used to obtain the UPDRS corresponding to other times by using segmented linear interpolation on the true UPDRS. Therefore, the UPDRS of patients at other times on this dataset was obtained by using piecewise linear interpolation on the true UPDRS. The values of motor-UPDRS and total-UPDRS range from 0-108 and 0-176, respectively. The higher the score, the more severe the symptoms of PD.

### B. Evaluation metrics

To evaluate the performance of the method in predicting PD severity, this paper uses mean absolute error (MAE), root mean square error (RMSE) and volatility (Vol) as the evaluation metrics of the method, calculated as follows:

$$MAE = \frac{1}{N_{test}} \sum_{i=1}^{N_{test}} |y_i - y'_i| \quad (16)$$

$$RMSE = \sqrt{\frac{1}{N_{test}} \sum_{i=1}^{N_{test}} (y'_i - y_i)^2} \quad (17)$$

$$Vol = \frac{\sum_{i=1}^{N_{test}} |v_i - \bar{v}|}{N_{test}} \quad (18)$$

where $N_{test}$ denotes the number of test instances, $y_i$ and $y'_i$ denote the true and predicted values of disease severity scores of PD patients, respectively. $v_i = y_i - y'_i$ represents the difference between the true value and the predicted value. $\bar{v} = \frac{1}{N_{test}} \sum_{i=1}^{N_{test}} v_i$ represents the mean of error fluctuations. MAE and RMSE indicate the difference between the predicted value of the model and the true value, and the smaller the value, the better the prediction performance of the model. Vol is used to assess the stability of the prediction model by quantifying the error fluctuations of the prediction results. The smaller the value, the more stable the prediction model is.

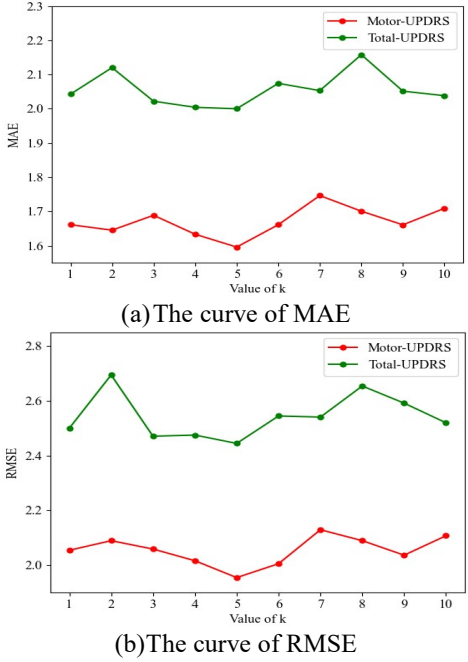

(a) The curve of MAE

(b) The curve of RMSE

**Fig.4.** Curves of MAE and RMSE with the number of transferred subjects.

TABLE II
MAIN PARAMETER SETTINGS OF PSGT METHOD

| Parameters | $k$ | $Spl_{num}$ | RF | |
|---|---|---|---|---|
| | | | $Num\_tree$ | $dep\_tree$ |
| Motor-UPDRS | 5 | $N/5$ | 30 | 50 |
| Total-UPDRS | 5 | $N/6$ | 50 | 50 |

*C. Experiment setup*

For experimental comparison, this paper treats each patient in the dataset as the target subject separately, and the other 41 patients as subjects in the source domain. The data of the target subject are randomly divided into training set, validation set and test set in the ratio of 6:2:2. Among them, the training set is used to fit the model, the validation set is used to analyze the similarity of diseases between the subjects and instances in the source domain and the target subject, and the test set is used to evaluate the effectiveness of the proposed prediction method. In the experimental process, the proposed PSGT method models each target patient separately, and takes the mean and standard deviation of the obtained MAE, RMSE and Vol to verify the feasibility and effectiveness of the PSGT method.

The main parameters of the experiments in this paper include the number $k$ of transferred subjects in the first stage of the PSGT method, the number $Spl_{num}$ of transferred instances in each subject in the second stage, and the number $Num\_tree$ and depth $dep\_tree$ of decision trees in the RF model.

The goal of the first stage of the proposed method in this paper is to identify subjects from the source domain that are similar to the target patient's disease. The value of $k$ in the first stage of the PSGT method determines the range of subjects to transfer. To determine the number $k$ of similar subjects, we visually show the graphs of the MAE and RMSE with the number $k$ of transferred subjects through Fig. 4. As shown in Fig. 4, the obtained MAE and RMSE errors are the smallest when $k$ is equal to 5. Therefore, the number $k$ of transferred subjects is set to 5 in this paper.

To avoid negative transfer, we have adjusted the number $Spl_{num}$ of transferred instances in the second stage to be smaller than the number $N$ of instances of the target patient. After testing, when predicting the motor-UPDRS and total-UPDRS of patients, we set the number $Spl_{num}$ of transferred instances to $N/5$ and $N/6$, respectively, resulting in better prediction performance. Additionally, we have tuned the two main parameters included in the RF model, i.e., the number of decision trees $Num\_tree$ and the maximum depth $dep\_tree$. Table II shows the specific settings of the main parameters in the PSGT method.

*D. Experiment results*

In this section, we conduct experiments on the Parkinson's telemonitoring dataset using traditional machine learning methods and existing methods proposed in the literature. We also compare the experiment results of the PSGT method with the above methods to verify the feasibility and effectiveness of the proposed method. In this paper, single prediction models (SVR, XGBoost, and RF), combined prediction methods (Boruta-XGBoost, FS-RF, and ARD-GPR), and transfer learning methods (ST, FDAMF, PTL, and TrAdaboost) are used for comparison and their parameters are optimized, respectively. Among them, Boruta-XGBoost [9], FS-RF [10], and ARD-GPR [11] are all PD severity prediction frameworks based on the combination of feature selection and machine learning. ST, FDAMF [5], PTL [4], and TrAdaboost [31] are transfer learning methods based on differences in disease progression among PD patients.

To ensure the fairness of the prediction experiments, all models have been trained and tested using the same training and testing sets, and the model parameters were optimized based on the validation set. Table III shows the mean and standard deviation of RMSE, MAE, and Vol for all prediction models.

As shown in Table III, a large prediction error is obtained in the single model using regressors (SVR, XGBoost, and RF) to predict the disease progression of PD patients, respectively. Compared with the single model, the combined model combines the feature selection method with the regressors, which can select the features with rich information, reduce the computational effort, and improve the performance of the model in predicting the severity of PD. In addition, the transfer learning methods ST, FDAMF, and PTL consider the similarity of data distribution between subjects, and obtain lower prediction errors. The TrAdaboost method also achieves better prediction performance by weighting the instances. However, relatively speaking, the predictive performance of the transfer learning model is worse than the combined model in the problem of predicting the severity of PD. This may be due to the difficulty in accurately measuring disease similarity between PD patients and the target subject when using transfer learning methods [5]. To improve the quality of transferred instances, the proposed PSGT method not only selects source subjects with similar disease development trends to the target patient, but also evaluates the contribution of instances in the transferred subjects. Therefore, the proposed PSGT method has the lowest mean and standard deviation of prediction error compared to the other prediction models. Specifically, the mean values of MAE, RMSE, and Vol obtained by predicting motor-UPDRS and total-UPDRS for target patients are 1.59, 1.95, 1.56 and 1.98, 2.54, 1.94, respectively. The results show that the PSGT method has superiority in both prediction error and stability, thus verifying that the method is effective and feasible in predicting the disease severity of PD patients.

TABLE III
COMPARISON OF PERFORMANCE OF DIFFERENT MODELS FOR PREDICTING PD SYMPTOM SEVERITY

| Methods | | Motor-UPDRS | | | Total-UPDRS | | |
|---|---|---|---|---|---|---|---|
| | | MAE | RMSE | Vol | MAE | RMSE | Vol |
| Single models | SVR | 6.66±3.47 | 7.20±3.43 | 2.29±1.17 | 8.32±5.01 | 8.98±4.86 | 2.66±0.94 |
| | XGBoost | 2.88±1.24 | 3.31±1.36 | 1.79±0.86 | 3.56±1.94 | 4.19±2.07 | 2.20±1.01 |
| | RF | 2.27±1.41 | 2.79±1.68 | 1.86±0.92 | 2.65±1.52 | 3.26±1.68 | 2.30±0.95 |
| Combined models | Boruta-XGBoost | 2.06±0.94 | 2.62±1.06 | 1.83±0.82 | 2.44±1.12 | 3.13±1.41 | 2.23±1.00 |
| | FS-RF | 1.97±0.96 | 2.53±1.26 | 1.84±0.89 | 2.38±1.16 | 2.98±1.46 | 2.23±0.99 |
| | ARD-GPR | 1.65±0.98 | 2.13±1.14 | 1.57±1.02 | 2.02±1.03 | 2.65±1.31 | 2.13±0.98 |
| Transfer learning models | ST | 1.70±0.90 | 2.08±1.09 | 1.68±0.89 | 2.10±1.08 | 2.58±1.30 | 2.08±1.07 |
| | FDAMF | 2.69±1.70 | 3.18±1.93 | 2.04±1.03 | 3.43±1.23 | 4.03±1.36 | 2.28±0.99 |
| | PTL | 2.11±1.33 | 2.51±1.56 | 1.89±1.09 | 2.50±1.04 | 2.89±1.22 | 2.17±1.03 |
| | TrAdaboost | 2.01±1.05 | 2.53±1.24 | 1.88±0.98 | 2.45±1.27 | 3.03±1.45 | 2.24±0.97 |
| Ours | PSGT | **1.59±0.88** | **1.95±1.03** | **1.56±0.86** | **1.98±0.91** | **2.54±1.21** | **1.94±0.89** |

## V. DISCUSSION

In this paper, an effective patient-specific game-based transfer method PSGT for PD severity prediction is proposed by analyzing the variability of disease among PD patients. The method selects subjects with small disease difference from the target patient by identifying the similarity of disease development trends between the source subjects and the target patient. The importance of the transferred subjects and their instances is also evaluated using the Shapley value, and a subset of instances with high contribution to the disease prediction of the target patient is selected. The experiment results show that the PSGT method outperforms the other compared methods in predicting the PD severity when the experiments are performed on the same Parkinson's

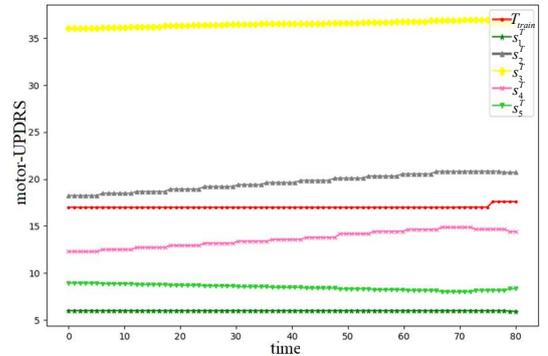

**Fig.5.** Comparison of the disease development trend of target subject $s_8$ and the transferred subjects over time for motor-UPDRS.

telemonitoring dataset, which verifies the feasible and effective of the PSGT method.

### A. Similarity between transferred subjects and the target patient

Instance transfer can effectively improve the accuracy of transfer learning method and facilitate the establishment of reliable prediction model by adding subject data from part of source domain to limited target data [16]. It can be seen from Table III that compared with the combined model and the transfer learning model, the error of the single model in predicting the disease severity of the target patient is larger. This may be because the single-model SVR, XGBoost, and RF methods directly transfer all PD patient data in the source domain to the target domain, without removing subjects that are significantly different from the target patient's disease development trend. This leads to the generation of negative transfer. Although the combined model does not filter the subjects in the source domain, it removes some redundant information through the feature selection method, which effectively reduces the prediction error of the model. The transfer learning methods are based on the difference of the disease development trend of PD patients, and aims to find data similar to the target patient's disease to improve the prediction performance of the model.

The PSGT method filters out the subjects with large differences from the target subject's disease development trend through the selection mechanism. Taking the PSGT method to predict the motor-UPDRS of target subject $s_8$ as an example, Fig. 5 shows the disease development trends over time between the five source domain subjects transferred by the model and the target subject $s_8$. As shown in Fig. 5, the subjects transferred by the PSGT method are very similar to the disease development trend of the target patient $s_8$. Therefore, the subject transfer in the PSGT method can select subjects with small disease differences from the target patient, thus laying a good foundation for transferring the effective subset of instances.

### B. Effectiveness of transferred instances

Inspired by cooperative game theory, this paper uses the Shapley value to assess the contribution of transferred subjects and instances. When performing the experiment on the Parkinson's telemonitoring dataset, a lower prediction error is obtained when the number of transferred subjects is 5. Therefore, the PSGT method uses the Shapley value to calculate the importance of the five transferred subjects, which can obtain the percentage of instances in each subject. Fig. 6 shows the proportion of transferred subjects that contain valid instances for predicting motor-UPDRS. As shown in Fig. 5, the overall variability of disease development trends between the transferred subjects and the target subject is small. However, as shown in Fig. 6, each of the transferred subjects contributes differently to the disease prediction of the target subject. This means that the proportion of instances with high contribution contained in the transferred subjects is different. If unimportant instances are added to the target subject data, it may reduce the performance of the disease severity prediction model. Therefore, it is necessary to select the subset of instances with higher contribution in the transferred subjects.

To highlight the validity of the subset of instances selected by the PSGT method, Table III compares the prediction errors of the subject transfer-based prediction method, i.e., ST and the PSGT method. The ST method adds all instances in the transferred subjects to the training set of the target patient without filtering out instances with less contribution. Therefore, compared with the prediction error of the ST method, the MAE, RMSE, and Vol obtained by the PSGT method predicting the motor-UPDRS and total-UPDRS are reduced by 6.47%, 6.25%, 7.14% and 5.71%, 1.55%, 6.73%, respectively. This experiment result indicates that the proposed PSGT further improves the performance of the method in predicting disease severity of PD patients by selecting instances with high contribution.

### C. Validating method interpretation

The PSGT uses the Shapley value to evaluate the importance of each transfer subject and its instances, and the subset of instances with high contribution is added to the target patient data to obtain a lower prediction error. The effectiveness of the Shapley value in evaluating the importance of subjects or instances is verified. In particular, the Shapley value enhances the interpretability of the PD severity prediction model by assessing the importance of each subject and instance.

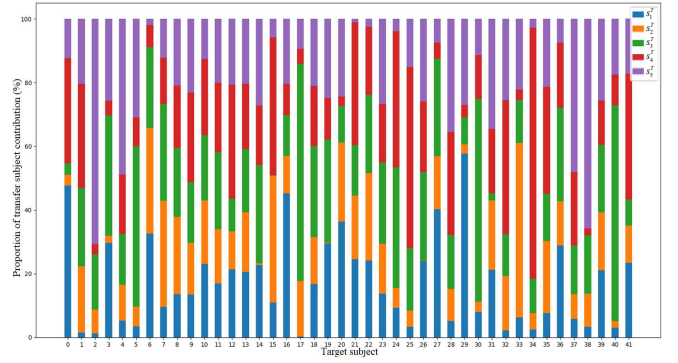

**Fig.6.** Contribution ratio of the five transferred subjects corresponding to each target subject for predicting motor-UPDRS.

Fig. 6 shows the contribution of each transferred subject to the disease prediction of the target patient for predicting motor-UPDRS. For example, for the target subject $s_8$ in Fig. 6, the Shapley value of each transferred subject is not much different, which indicates that the contribution of the transferred subject data to the disease prediction of the target subject $s_8$ is basically the same. As shown in Fig. 5, the five transferred subjects have strong similarity with the disease development trend of target subject $s_8$. This is consistent with the calculation results of the Shapley value, which verifies the effectiveness of the Shapley value for enhancing the interpretability of the model. However, for target subject $s_2$ in Fig. 6, the Shapley values of transferred subjects $s_3^T$ and $s_5^T$ are large, while the Shapley values of $s_1^T$, $s_2^T$ and $s_4^T$ are very small. This implies that most of the instances with high contribution to the prediction of disease in the target patient are in $s_3^T$ and $s_5^T$. The reason for this phenomenon may be that in the Parkinson's telemonitoring dataset, the number of subjects with less disease difference from the target patient $s_2$ is small. Therefore, the contribution of some transferred subjects to the disease prediction of the target patient is very small when calculated using the Shapley value.

### D. Comparative analysis of transfer learning methods

In the transfer learning methods for PD severity prediction, the proposed PSGT method obtains the lowest MAE, RMSE, and Vol, thus demonstrating its good prediction performance. The specific analysis of this experiment result is as follows.

The ST method selects subjects from the source subjects with similar disease development trends to the target subject, which compensates for the lack of training data for the target patient. The FDAMF method [5] uses a filtering mechanism to select subjects with similar data distribution to the target subject, and then uses the selected subjects to train suitable ridge regression model parameters for disease prediction of the target subject. Similarly, the PTL method [4] evaluates the risk of negative transfer for each source subject and determines the specific model parameters to be transmitted to the target patient by the selected subjects, which improves the prediction performance of the model and had a very strong robust to negative transfer. The above approach is to find subjects from the source domain that are similar to the disease development trend of the target patient, and then train the model parameters for transfer based on the selected subjects. However, as shown in Fig. 6, the proportion of instances with high contributions that are contained within the transferred subjects is not the same. Since the parameters of the models trained by FDAMF and PTL methods are not accurate enough, the performance of the prediction model is affected. The TrAdaBoost algorithm can select instances similar to the target patient's disease progression through the automatic weight update mechanism. However, in the Parkinson's telemonitoring dataset, the instance distributions in the source and target domains may be very different, which makes the TrAdaBoost algorithm prone to overfitting and negative transfer problems, resulting in the degradation of model performance [48], [49]. The proposed PSGT method selects subjects with small differences from the target patient's disease through subject transfer, and the risk of negative transfer is further reduced by the contribution evaluation and selection of the instances in the transferred subjects. Therefore, the PSGT method selects a subset of instances with high contribution to the disease prediction of the target patient through the game transfer mechanism, and has better performance in both prediction error and stability among the compared methods, thus verifying the effectiveness of the proposed method.

*E. Limitations of the method*

Although the PSGT can improve the accuracy of the PD severity prediction model, it still has some limitations. During the subject transfer process, we fixedly transfer five PD patients for each target patient from a holistic perspective. However, the number of source subjects with similar disease development trends to each target subject in the dataset may not be the same. Therefore, how to select the number of transferred subjects adaptively needs further study. In addition, with the development of multimodal-based PD severity analysis [50], it will be considered to transfer instances of other modalities such as patient's voice, gait, and tapping to the target domain to further improve the model performance.

VI. CONCLUSION

In this paper, a patient-specific game-based transfer method PSGT for PD severity prediction is proposed by analyzing the difference of the severity of PD patients. The method can select subjects with similar disease development trends to the target subject from the source domain, and fairly evaluate the importance of transferred subjects and their instances by the Shapley value, which also enhances the interpretability of the method. The problem of insufficient patient training data in the transfer method is effectively alleviated by adding the subset of instances with high contribution to the dataset of the target subject. Finally, the prediction performance of the method is improved by using the RF model to predict the disease severity of the target patient. In this paper, the proposed PSGT method is tested on the Parkinson's telemonitoring dataset. The results show that the PSGT exhibits better performance in terms of prediction error and stability compared to existing single models, combined models, and transfer learning methods for PD severity, which proves the feasibility and effectiveness of the proposed method.

In future research, considering that the number of source subjects with similar disease development trends to the target subject may not be the same, a method for adaptively selecting the number of transferred subjects will be designed to reduce the risk of negative transfer. And, the multimodal instance transfer framework will be further studied, which also make the research more challenging and interesting.